\documentclass[letterpaper]{article} % DO NOT CHANGE THIS
\usepackage[preprint]{aaai2027} % DO NOT CHANGE THIS
\usepackage[hyphens]{url} % DO NOT CHANGE THIS
\usepackage{graphicx} % DO NOT CHANGE THIS
\usepackage{natbib} % DO NOT CHANGE THIS AND DO NOT ADD OPTIONS
\usepackage{caption} % DO NOT CHANGE THIS AND DO NOT ADD OPTIONS
\usepackage{amsmath}
\usepackage{amssymb}
\usepackage{booktabs}
\usepackage{array}
\usepackage{algorithm}
\usepackage{algorithmic}
\usepackage[most]{tcolorbox}
\usepackage{fvextra}

\newcommand{\conrub}{\textsc{ConRub-Med}}
\newcommand{\grpo}{\textsc{GRPO}}
\newcommand{\correct}{\textsc{correct}}
\newcommand{\missing}{\textsc{missing}}
\newcommand{\wrong}{\textsc{wrong}}
\newcommand{\closedref}[1]{\textcolor[gray]{0.45}{#1}}
\newcolumntype{P}[1]{>{\raggedright\arraybackslash}p{#1}}

\definecolor{SuppBlack}{HTML}{202020}
\definecolor{SuppDarkGray}{HTML}{555555}
\definecolor{SuppLightGray}{HTML}{EEEEEE}
\definecolor{SuppPaperGray}{HTML}{F7F7F7}
\definecolor{SuppPanelGray}{HTML}{F0F0F0}
\definecolor{SuppCodeGray}{HTML}{F7F7F7}

\newtcolorbox{PromptFrame}[1]{%
  enhanced,
  breakable,
  colback=SuppCodeGray,
  colframe=SuppDarkGray,
  colbacktitle=SuppBlack,
  coltitle=white,
  fonttitle=\bfseries\small,
  title={#1},
  title after break={#1\space (continued)},
  boxrule=0.45pt,
  arc=1.2mm,
  left=1.6mm,
  right=1.6mm,
  top=1.2mm,
  bottom=1.2mm,
  before skip=5pt,
  after skip=8pt
}

\DefineVerbatimEnvironment{PromptText}{Verbatim}{%
  fontsize=\footnotesize,
  breaklines=true,
  breakanywhere=true,
  breakautoindent=false,
  breakindent=0pt,
  breaksymbolleft={},
  breaksymbolright={},
  breakanywheresymbolpre={},
  breakanywheresymbolpost={}
}

\title{ConRub-Med: Reinforcement Learning with Consensus Rubrics for Open-Ended Medical Question Answering}
\author{
Taojie Zhu\textsuperscript{\rm 1,\rm 2}\thanks{This work was completed during an internship at Ant Group.},
Yuan Xia\textsuperscript{\rm 2}\corresponding,
Tao Sun\textsuperscript{\rm 2},
Yizhi Wang\textsuperscript{\rm 1},
Yan Chen\textsuperscript{\rm 1}\\
Qunshan He\textsuperscript{\rm 2,\rm 3},
Tian Guan\textsuperscript{\rm 1},
Jian Wang\textsuperscript{\rm 2},
Jinjie Gu\textsuperscript{\rm 2},
Junwei Liu\textsuperscript{\rm 2},
Yonghong He\textsuperscript{\rm 1}\corresponding
}
\affiliations{
\textsuperscript{\rm 1}Tsinghua University \quad
\textsuperscript{\rm 2}Ant Group \quad
\textsuperscript{\rm 3}Zhejiang University\\
zhongjing.xy@antgroup.com, heyh@sz.tsinghua.edu.cn
}

\begin{document}

\maketitle

\begin{abstract}
Reinforcement learning with verifiable rewards has been especially effective in mathematics and coding, where answers can be checked automatically. Many open-ended medical questions lack comparably cheap outcome verifiers: responses may be partly correct, incomplete, or contain clinically consequential errors. Rubrics written or validated by physicians offer strong clinical grounding, but involving experts in every instance is costly. Model-generated rubrics make this supervision scalable. We introduce \conrub{} to preserve useful distinctions as rubric feedback moves from construction to policy optimization. For each prompt, three heterogeneous language models propose atomic criteria independently; a separate model reviews them, retaining only criteria with semantic support from all three generators. Three-State scoring distinguishes correct coverage, missing information, and incorrect claims. Errors receive negative rather than zero credit. When every response in a complete Group Relative Policy Optimization (GRPO) group receives the same final reward, a pairwise judge provides sequence advantages only if both candidate orders agree, without changing the scalar rewards. Groups without ties use vanilla GRPO. In a blinded study matched by question, two medical experts rate panels from the full pipeline as more clinically relevant than panels produced by one generator. Across the evaluated open models, \conrub{} ranks first on six of nine benchmarks and achieves the highest medical and generalization averages. Using the resulting rubric dataset of 5,166 prompts, it scores \(38.98\pm1.04\) (mean \(\pm\) SD) on HealthBench-Hard, compared with InfiMed-ORBIT's 33.60 with 8,000 samples and 37.30 with 28,000.
\end{abstract}

\section{Introduction}

Reinforcement learning with verifiable rewards (RLVR) has made large-scale policy exploration effective most visibly in mathematics and software reasoning. These domains offer relatively cheap, repeatable checks, such as answer matching or rule-based validation, so many sampled trajectories can be evaluated without human labels for every rollout~\citep{shao2024deepseekmathpushinglimitsmathematical,wei2025swerl}. This scaling depends on an automatic outcome signal that can score every sampled trajectory.

Many open-ended medical questions lack a comparably cheap, single outcome verifier. Their answers are not simply right or wrong: they may express the same content in different forms, be partly correct yet incomplete, or contain one unsafe claim amid otherwise useful content. These distinctions are especially consequential in medical QA, where quality spans factuality, relevance, safety, and appropriate uncertainty~\citep{arora2025healthbenchevaluatinglargelanguage}. The challenge is therefore not merely to sample more responses, but to obtain feedback rich enough to identify medically meaningful differences among them.

Rubrics written or validated by physicians offer strong clinical grounding, but involving experts in every instance is costly and hard to scale~\citep{arora2025healthbenchevaluatinglargelanguage,lyu2026clinalign}. Rubrics generated by models offer a scalable alternative, and several recent methods use them in place of a single outcome check. Rubrics as Rewards uses a rubric tailored to each prompt as an on-policy reward signal for medical and scientific reasoning, while QuRL constructs rubrics for individual open-ended QA cases~\citep{gunjal2026rubricsasrewards,wei2026qurl}. InfiMed-ORBIT constructs a rubric for each medical dialogue and masks groups with insufficient reward spread~\citep{wang2026infimedorbit}. RubricHub pools criteria from different generators, whereas OpenRubrics uses preferences to guide rubric synthesis~\citep{li2026rubrichub,liu2026openrubrics}. Their usefulness depends on how criteria are constructed and whether scoring and optimization preserve meaningful differences among responses. RubricBench reports a substantial gap between rubrics annotated by experts and those generated by models~\citep{zhang2026rubricbench}. Separately, if criterion judgments aggregate to identical final rewards, vanilla group-relative optimization provides no direction within the group~\citep{shao2024deepseekmathpushinglimitsmathematical}.

We introduce \conrub{} to strengthen the path from rubric construction to policy optimization. First, three models generate criteria independently. A separate reviewer filters them and retains only criteria supported by all three. This process also yields a new rubric training dataset. Second, Three-State scoring asks the criterion judge to distinguish correct, missing, and wrong content, assigning negative credit to errors. Third, when the final rewards in a complete \grpo{} group are finite and identical, a separate judge compares response pairs in both orders. Preferences that agree across the two orders become Pairwise Sequence Advantages.

\begin{figure*}[t]
\centering
\includegraphics[width=0.97\textwidth]{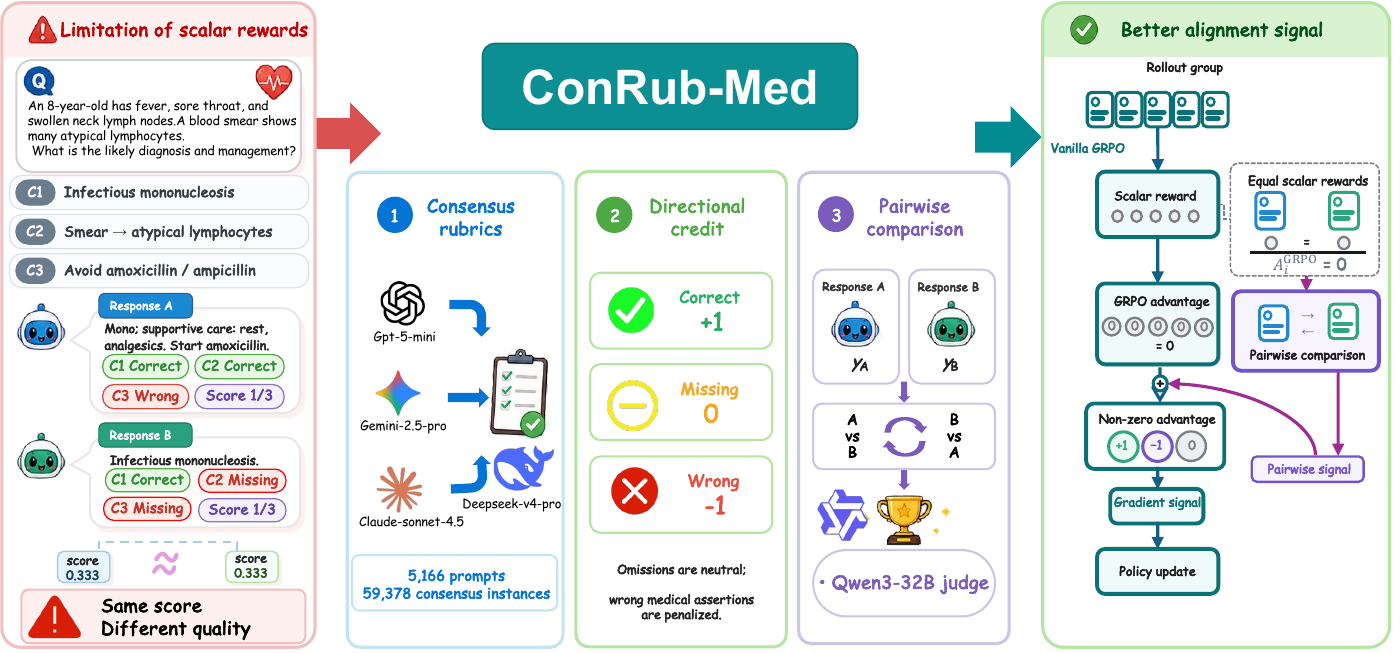}
\caption{Overview of \conrub{}. The left panel shows how criterion profiles \((+1,+1,-1)\) and \((+1,0,0)\) can receive the same mean score of \(1/3\). The remaining panels summarize consensus rubric construction, three-state scoring, and pairwise sequence advantages for exact final-reward ties.}
\label{fig:conrub-overview}
\end{figure*}

Our main contributions are threefold.
\textbf{(i)} We introduce Consensus Rubric Construction, in which three models generate criteria independently and a separate reviewer retains criteria supported by all three. This process produces a training set of 5,166 prompts with 49,046 content criteria and 10,332 global controls. We will publicly release the dataset upon paper acceptance.
\textbf{(ii)} We develop a training procedure that connects criterion scoring with policy optimization. Three-State Criterion Scoring distinguishes correct, missing, and wrong content. For complete \grpo{} groups whose final rewards are finite and identical, Pairwise Sequence Advantages use comparisons that agree across both response orders and write them directly as sequence advantages, without adding a separate preference loss.
\textbf{(iii)} We evaluate \conrub{} through an expert study blinded to criterion source and matched by question, controlled comparisons of its components, and nine benchmarks. Both medical experts give higher clinical relevance ratings to criteria from the complete construction pipeline than to criteria produced by one generator. Among the evaluated models with open weights, \conrub{} ranks first on six of nine benchmarks, obtains the highest medical and generalization averages, and scores 38.98 on HealthBench-Hard.

\begin{figure*}[t]
\centering
\includegraphics[width=0.97\textwidth]{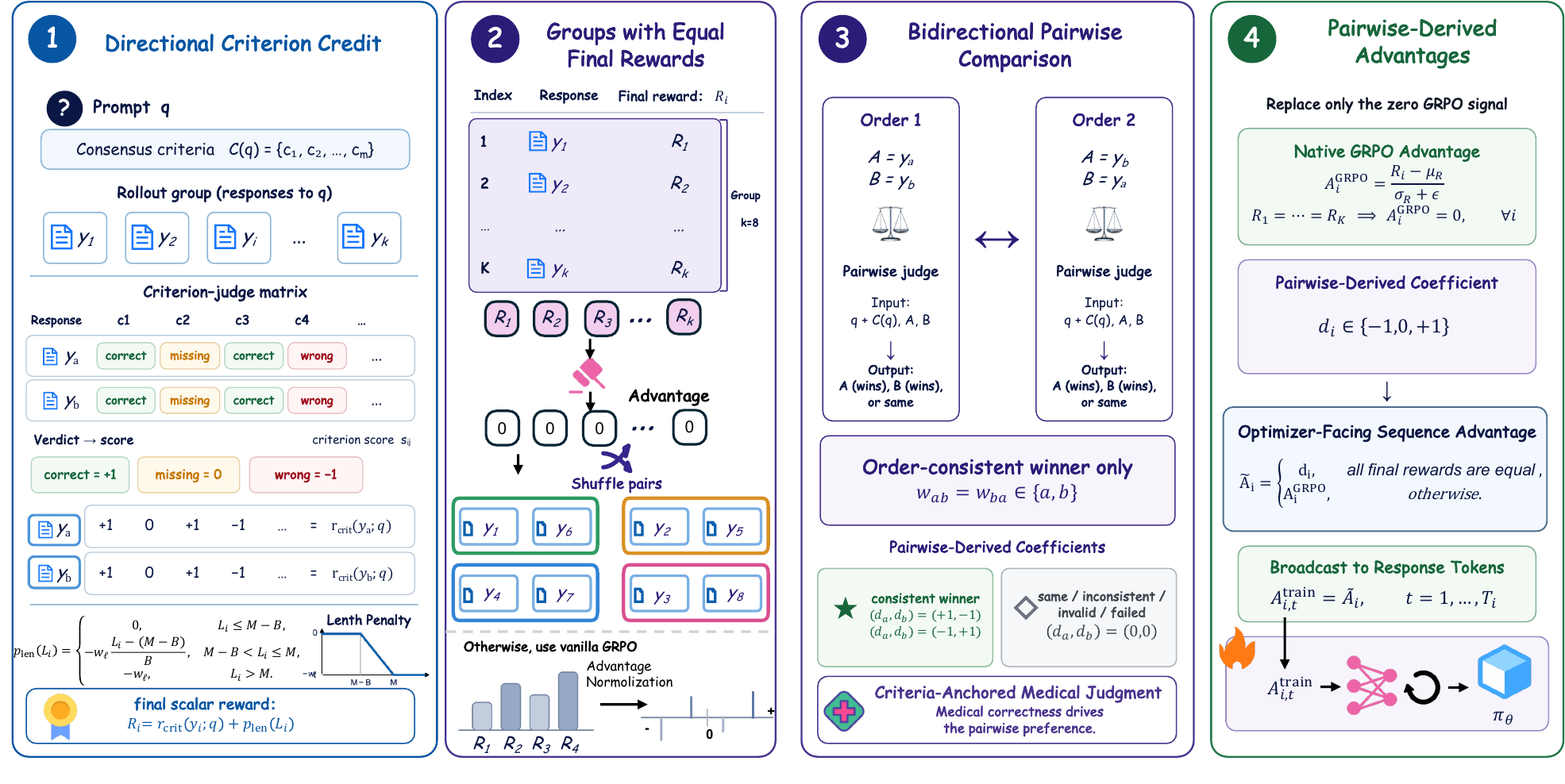}
\caption{Criterion-to-advantage optimization. Three-state scores and a soft length penalty define scalar rewards. Complete eight-response groups with identical finite rewards are divided into four pairs and judged in both orders. Order-consistent winners receive \((+1,-1)\) sequence advantages; other pairwise outcomes yield zero. Non-tied groups retain vanilla \grpo{} advantages.}
\label{fig:preference-advantages}
\end{figure*}

\section{Related Work}

\subsection{Rubric Supervision for Open-Ended Medical QA}

Open-ended medical QA requires feedback that captures factuality, completeness, safety, and appropriate uncertainty beyond exact answer matching. HealthBench addresses this need with criteria written by physicians for each prompt, while ClinAlign distills rubrics refined by clinicians into reusable health principles~\citep{arora2025healthbenchevaluatinglargelanguage,lyu2026clinalign}. RubricHub builds rubrics through successive coarse and fine stages, whereas OpenRubrics uses preferences to guide rubric synthesis~\citep{li2026rubrichub,liu2026openrubrics}. Yet RubricBench reports a persistent gap between rubrics annotated by experts and those generated by models~\citep{zhang2026rubricbench}. \conrub{} combines independent generation, model review, and cross-model semantic agreement to reduce reliance on any single generator.

\subsection{Rubric-Based Reinforcement Learning}

Rubric-based RL turns prompt-specific criteria into structured feedback. Rubrics as Rewards (RaR) formalizes binary criterion rewards, QuRL mines case-wise rubrics, and RubricHub combines rubric-guided rejection sampling with RL~\citep{gunjal2026rubricsasrewards,wei2026qurl,li2026rubrichub}. RaR compares weighted aggregation with holistic scoring, while alternating reinforcement learning with contextual rubric rewards (ARL-RR) alternates across semantic rubric classes to avoid fixed scalarization~\citep{gunjal2026rubricsasrewards,lan2026alternatingreinforcementlearningcontextual}. InfiMed-ORBIT generates weighted positive and negative criteria, which are judged as satisfied or not~\citep{wang2026infimedorbit}. \conrub{} takes a simpler approach: it uses equally weighted positive criteria and leaves the distinction between \correct{}, \missing{}, and \wrong{} to the judge. The rubric generator only needs to describe what a good answer should contain, without assigning importance weights or anticipating specific errors.

\subsection{Pairwise Feedback for Exact Reward Ties}

Pairwise preferences classically supervise scalar reward models for reinforcement learning from human feedback (RLHF), as in preference-based RL and InstructGPT~\citep{christiano2017deep,ouyang2022instructgpt}. DPO directly optimizes an offline preference objective, whereas P3O uses online pairwise reward differences in policy updates~\citep{rafailov2023dpo,wu2024p3o}. In group-relative optimization, \grpo{} derives advantages from within-group scalar rewards; DAPO filters homogeneous groups, and InfiMed-ORBIT masks groups with insufficient reward spread~\citep{shao2024deepseekmathpushinglimitsmathematical,yu2025dapo,wang2026infimedorbit}. Lanchantin et al. form verifier-defined correct--incorrect pairs for a GroupDPO loss and separately combine it with \grpo{}; DYPO uses reward-ordered mixed groups, and AMIR-GRPO uses margin-separated within-group reward rankings~\citep{lanchantin2025bridgingofflineonlinereinforcement,zhu2026dypo,yari2026amirgrpoinducingimplicitpreference}. Unlike these broadly applied comparative objectives, \conrub{} invokes a separately prompted pairwise judge only for complete groups with exactly equal final scalar rewards. Each pair is evaluated in both candidate orders, and order-consistent preferences are written directly into sequence advantages.

\section{Method}
\label{sec:method}

\subsection{Overview}

Let \(q\) be an open-ended prompt and \(C(q)=\{c_1,\ldots,c_m\}\) its criterion set. A policy \(\pi_\theta\) samples a response group \(Y=(y_1,\ldots,y_k)\), with \(k=8\), and a judge returns one verdict \(v_{ij}\) for each response--criterion pair. Figures~\ref{fig:conrub-overview} and~\ref{fig:preference-advantages} summarize the method: consensus construction defines \(C(q)\), three-state scoring maps criterion verdicts to a scalar reward \(R_i\), and bidirectionally consistent comparisons supply pair-local sequence advantages when a complete group has exactly equal final rewards.

\subsection{Consensus Rubric Construction}

If a generated criterion is incorrect, every sampled response for that prompt is judged against it. To reduce dependence on any one generator, GPT-5 Mini~\citep{openai2025gpt5mini}, Gemini 2.5 Pro~\citep{comanici2025gemini25pushingfrontier}, and Claude Sonnet 4.5~\citep{anthropic2025claudesonnet45} independently produce 20--30 candidate criteria from the question and available reference context. Each candidate is written as an atomic requirement that can be judged independently. A reference-coverage tag records whether the reference addresses the candidate; it is used during construction rather than as a reward label.

Deterministic checks remove empty, underspecified, or internally repetitive outputs. DeepSeek-V4-Pro (\texttt{deepseek-v4-pro})~\citep{deepseekai2026deepseekv4highlyefficientmilliontoken} then screens the remaining criteria for domain errors, logical contradictions, and reference conflicts, and clusters criteria that express the same underlying proposition. Let \(C^{(g)}(q)\) be the candidates from generator \(g\) and \(G_\ell\) a semantic cluster. Its support set is \(S_\ell=\{g:C^{(g)}(q)\cap G_\ell\neq\varnothing\}\). We retain \(G_\ell\) only when \(|S_\ell|=3\) and, where reference-coverage tags are available, the tags agree within the cluster.

Each retained cluster contributes one final criterion. Its longest member is used as the initial representative; the reviewer either keeps it, rewrites it as a clear positive requirement, or excludes it. After removing exact duplicates, we retain at most ten content criteria under fixed dimension quotas and append two global controls for factual correctness and response appropriateness. Both content criteria and global controls enter three-state scoring and pairwise judging.

The final training set contains 5,166 prompts and 59,378 criterion instances, including 49,046 prompt-specific criteria and 10,332 global controls. The supplementary material gives the prompts, thresholds, source allocation, and complete construction statistics.

\subsection{Criterion-to-Advantage Optimization}

\paragraph{Binary and three-state scoring.}
The binary baseline asks whether each criterion is satisfied:
\begin{equation}
\begin{aligned}
b_{ij}&\in\{\texttt{yes},\texttt{no}\},\\
r_{\mathrm{bin}}(y_i;q)
&=\frac{1}{|C(q)|}\sum_{j=1}^{|C(q)|}
\mathbb{1}[b_{ij}=\texttt{yes}].
\end{aligned}
\end{equation}
Because missing and incorrect content both receive \texttt{no}, this reward cannot distinguish omission from an incorrect assertion about the same criterion.

The three-state judge instead returns
\begin{equation}
\begin{aligned}
v_{ij}&\in\{\correct,\missing,\wrong\},\\
s_{ij}&=
\begin{cases}
+1,&v_{ij}=\correct,\\
0,&v_{ij}=\missing,\\
-1,&v_{ij}=\wrong.
\end{cases}
\end{aligned}
\end{equation}
\wrong{} is reserved for a factual error, fabrication, contradiction, reversal, or substitution concerning the criterion; imprecision alone is not automatically wrong. We define
\begin{equation}
r_{\mathrm{crit}}(y_i;q)
=\frac{1}{|C(q)|}\sum_{j=1}^{|C(q)|}s_{ij}.
\label{eq:criterion-reward}
\end{equation}
Both pointwise variants add the same soft-overlength penalty~\citep{yu2025dapo}:
\[
R_i=r(y_i;q)+p_{\mathrm{len}}(L_i),
\qquad r\in\{r_{\mathrm{bin}},r_{\mathrm{crit}}\}.
\]
Pairwise eligibility is determined from this final reward. The supplementary material specifies the penalty.

\paragraph{\grpo{}-based policy optimization.}

For a complete group, group-relative reward normalization forms the sequence advantage~\citep{shao2024deepseekmathpushinglimitsmathematical}
\begin{equation}
\mu_R=\frac{1}{k}\sum_{i=1}^{k}R_i,\qquad
A_i^{\mathrm{GRPO}}=\frac{R_i-\mu_R}{\sigma_R+\varepsilon},
\label{eq:vanilla-grpo}
\end{equation}
where \(\sigma_R\) is the within-group sample standard deviation and \(\varepsilon=10^{-6}\). For token \(t\) of response \(i\), define the current-to-old-policy importance ratio
\[
\rho_{i,t}(\theta)=
\frac{\pi_\theta(y_{i,t}\mid q,y_{i,<t})}
{\pi_{\theta_{\mathrm{old}}}(y_{i,t}\mid q,y_{i,<t})}.
\]
Let \(A_{i,t}^{\mathrm{train}}\) denote the advantage supplied to that token. All variants use the following per-group objective, combining group-relative advantages with asymmetric clipping~\citep{yu2025dapo}:
\begin{equation}
\begin{aligned}
\mathcal L_{\mathrm{GRPO}}(\theta)
&=-\frac{1}{k}\sum_{i=1}^{k}\frac{1}{T_i}
\sum_{t=1}^{T_i}
\min\Bigl(
\rho_{i,t}(\theta)A_{i,t}^{\mathrm{train}},\\[-2pt]
&\quad\operatorname{clip}\!\left(
\rho_{i,t}(\theta),
1-\epsilon_{\mathrm{low}},
1+\epsilon_{\mathrm{high}}
\right)A_{i,t}^{\mathrm{train}}
\Bigr).
\end{aligned}
\label{eq:grpo-objective}
\end{equation}
where \(T_i\) is the response length, and \(\epsilon_{\mathrm{low}}\) and \(\epsilon_{\mathrm{high}}\) are the lower and upper clipping widths. Without pairwise intervention, \(A_{i,t}^{\mathrm{train}}=A_i^{\mathrm{GRPO}}\).

\paragraph{Pairwise advantages for exact reward ties.}

We apply pairwise comparison only to complete eight-response groups with no removed or judge-failed response and with finite, exactly equal final rewards:
\begin{equation}
R_1=\cdots=R_k,
\label{eq:equal-final-rewards}
\end{equation}
which implies \(A_i^{\mathrm{GRPO}}=0\) for every response. A seeded shuffle partitions the eight responses into four disjoint pairs, so each response is compared once.

For each pair \((y_a,y_b)\), the pairwise judge receives the question, the full criterion set, and both responses, but not their rewards or the reference answer. It selects the better overall response, using medical correctness as its primary basis. Clear medical errors, unsafe claims, or contradictions outside the listed criteria may distinguish the responses, whereas uncertain claims outside the criteria do not break a tie. To control for position bias in model-based evaluation~\citep{wang2024unfairevaluators}, we compare both candidate orders. Let \(w_{ab}\) and \(w_{ba}\) denote outcomes mapped back to response identity under the original and swapped orders. We assign
\[
(d_a,d_b)=
\begin{cases}
(+1,-1),&w_{ab}=w_{ba}=a,\\
(-1,+1),&w_{ab}=w_{ba}=b,\\
(0,0),&\text{otherwise}.
\end{cases}
\]
Only an order-consistent preference is nonzero. We do not renormalize these coefficients, so groups with fewer consistent pairs carry less pairwise signal. Collecting pair assignments gives \(d_i\in\{-1,0,+1\}\), and the optimizer-facing sequence advantage is
\begin{equation}
\boxed{
\widetilde{A}_i=
\begin{cases}
d_i,&\text{if all final rewards are equal},\\
A_i^{\mathrm{GRPO}},&\text{otherwise}.
\end{cases}
}
\label{eq:preference-advantage}
\end{equation}
For the full method, \(\widetilde A_i\) is broadcast as \(A_{i,t}^{\mathrm{train}}=\widetilde A_i\) for \(t=1,\ldots,T_i\). Scalar rewards remain unchanged, all other groups retain the same group-relative advantage, and no separate preference loss is added. The supplementary material gives the complete procedure and defines the trace metrics.

\begin{table*}[!t]
\centering
\small
\setlength{\tabcolsep}{1.5pt}
\begin{tabular}{@{}lccccccccccc@{}}
\toprule
& \multicolumn{7}{c}{Medical Evaluation} & \multicolumn{4}{c}{Generalization \& Retention} \\
\cmidrule(lr){2-8}\cmidrule(lr){9-12}
Model / Method & HB-Hard & MedXpert & \shortstack{Diag.\\Arena} & \shortstack{Med\\MCQA} & \shortstack{PubMed\\QA} & \shortstack{MMLU\\Med} & \underline{Med.\ Avg} & Writing & \shortstack{GPQA\\D} & \shortstack{IFEval\\L} & \underline{Gen.\ Avg} \\
\midrule
\multicolumn{12}{@{}l}{\emph{Open-Weight}} \\[-1pt]
MedGemma-1.5-4B-IT & 0.00 & 9.96 & 35.60 & 41.38 & 64.00 & 65.69 & 36.11 & 43.07 & 10.91 & 48.43 & 34.14 \\
HuatuoGPT-o1-7B & 0.00 & 14.53 & 39.40 & \textbf{62.97} & \underline{74.20} & 74.36 & 44.24 & 47.70 & 20.71 & 55.45 & 41.29 \\
\addlinespace[1pt]
\multicolumn{12}{@{}l}{\emph{Closed-Source Models (Used for Rubric Construction)}} \\[-1pt]
\closedref{GPT-5 Mini} & \closedref{37.74} & \closedref{45.14} & \closedref{65.60} & \closedref{80.61} & \closedref{73.80} & \closedref{89.74} & \closedref{65.44} & \closedref{84.09} & \closedref{76.97} & \closedref{90.76} & \closedref{83.94} \\
\closedref{Gemini 2.5 Pro} & \closedref{15.97} & \closedref{46.24} & \closedref{68.50} & \closedref{84.13} & \closedref{73.60} & \closedref{90.48} & \closedref{63.15} & \closedref{79.93} & \closedref{53.74} & \closedref{93.72} & \closedref{75.80} \\
\closedref{Claude Sonnet 4.5} & \closedref{10.62} & \closedref{37.18} & \closedref{65.50} & \closedref{79.78} & \closedref{73.40} & \closedref{91.30} & \closedref{59.63} & \closedref{53.59} & \closedref{70.71} & \closedref{91.50} & \closedref{71.93} \\
\midrule
Qwen3-4B-Instruct-2507 & 6.30 & \underline{16.57} & \underline{40.70} & 57.52 & \textbf{76.40} & \underline{78.36} & \underline{45.98} & 82.55 & 41.98 & \textbf{87.25} & \underline{70.59} \\
\textsc{GRPO + DPO} & \underline{17.47} & 11.92 & 36.90 & 53.69 & 60.00 & 78.13 & 43.02 & \underline{86.44} & \underline{45.15} & \underline{77.45} & 69.68 \\
\conrub{} (Ours) & \textbf{38.98} & \textbf{16.98} & \textbf{44.20} & \underline{59.12} & 71.80 & \textbf{79.49} & \textbf{51.76} & \textbf{88.75} & \textbf{59.39} & 74.68 & \textbf{74.27} \\
\bottomrule
\end{tabular}
\caption{Main results. Med.\ Avg and Gen.\ Avg are unweighted macro-averages over their respective benchmark groups. Gray rows show closed-source references; bold and underlined values denote the best and second-best open-weight scores in each column.}
\label{tab:main-results}
\end{table*}

\section{Experiments}

\begin{figure}[t]
\centering
\includegraphics[width=\columnwidth]{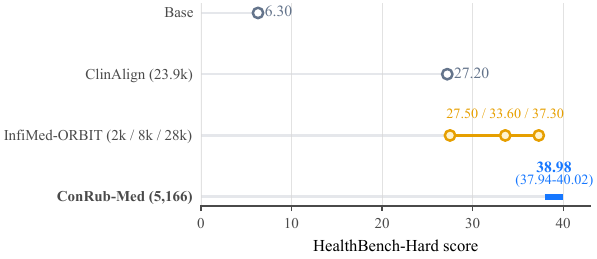}
\caption{HealthBench-Hard literature comparison for Qwen3-4B-Instruct-family policies evaluated with the official GPT-4.1 judge. Parentheses indicate the number of training prompts; the \conrub{} marker shows the mean and range across three evaluation seeds.}
\label{fig:healthbench-hard-literature}
\end{figure}

\setcounter{dbltopnumber}{3}

\begin{table*}[!t]
\centering
\small
\setlength{\tabcolsep}{1.5pt}
\begin{tabular}{@{}lccccccccccc@{}}
\toprule
& \multicolumn{7}{c}{Medical Evaluation} & \multicolumn{4}{c}{Generalization \& Retention} \\
\cmidrule(lr){2-8}\cmidrule(lr){9-12}
Variant & HB-Hard & MedXpert & \shortstack{Diag.\\Arena} & \shortstack{Med\\MCQA} & \shortstack{PubMed\\QA} & \shortstack{MMLU\\Med} & \underline{Med.\ Avg} & Writing & \shortstack{GPQA\\D} & \shortstack{IFEval\\L} & \underline{Gen.\ Avg} \\
\midrule
Qwen3-4B-2507 (Base) & 6.30 & \underline{16.57} & 40.70 & 57.52 & \textbf{76.40} & 78.36 & 45.98 & 82.55 & 41.98 & \textbf{87.25} & 70.59 \\
\midrule
\multicolumn{12}{@{}l}{\textbf{+ Single-generator rubrics (one source per run)}} \\
\quad GPT-5 Mini & 26.02 & 16.37 & 41.00 & 56.80 & 71.60 & \underline{81.50} & 48.88 & 88.42 & 48.28 & 82.26 & 72.99 \\
\quad Gemini 2.5 Pro & 23.56 & 15.59 & \underline{45.00} & 55.70 & 72.20 & \textbf{81.53} & 48.93 & 88.19 & 49.39 & \underline{82.81} & \underline{73.46} \\
\quad Claude Sonnet 4.5 & 23.99 & 14.29 & 37.50 & \underline{58.02} & \underline{72.60} & 78.64 & 47.51 & 86.49 & 48.08 & 80.59 & 71.72 \\
+ Consensus Rubrics & 32.71 & 15.10 & 43.80 & 57.97 & 64.80 & 81.18 & 49.26 & \textbf{90.69} & \underline{58.08} & 45.29 & 64.69 \\
\midrule
\quad + Three-State & \underline{36.40} & 15.76 & \textbf{47.10} & 56.97 & 71.60 & 81.23 & \underline{51.51} & 88.65 & 55.86 & 44.36 & 62.96 \\
\quad + Pairwise (\conrub{}) & \textbf{38.98} & \textbf{16.98} & 44.20 & \textbf{59.12} & 71.80 & 79.49 & \textbf{51.76} & \underline{88.75} & \textbf{59.39} & 74.68 & \textbf{74.27} \\
\bottomrule
\end{tabular}
\caption{Ablations of rubric source, scoring, and tie handling. Rubric-source variants use binary scoring on matched 5,166-prompt sets; indented rows retain Consensus Rubrics and vary scoring or tie handling. Med.\ Avg and Gen.\ Avg are unweighted macro-averages. Bold and underlined values denote the best and second-best scores in each column.}
\label{tab:ablations}
\end{table*}

\begin{figure*}[!t]
\centering
\includegraphics[width=0.97\textwidth]{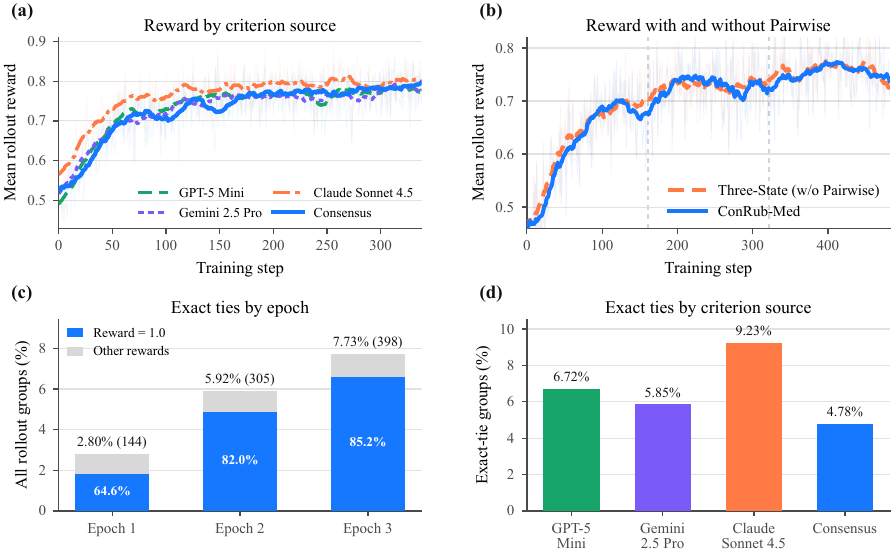}
\caption{Training rewards and exact ties. (a) Criterion-source rewards over the first approximately 340 steps. (b) Three-State rewards with and without Pairwise Advantages. (c) Exact final-reward ties by epoch; blue denotes reward 1.0, gray all other rewards, and in-bar percentages give the share of exact ties at reward 1.0. (d) Exact-tie rates for the matched binary criterion-source runs.}
\label{fig:training-reward-trajectories}
\end{figure*}

\subsection{Experimental Setup}

\paragraph{Models and baselines.}
All controlled variants initialize from Qwen3-4B-Instruct-2507~\citep{qwen2025qwen34binstruct2507}, with Qwen3-32B~\citep{yang2025qwen3technicalreport} serving as both the criterion judge and, under a separate prompt, the pairwise judge. Criterion-source ablations use GPT-5 Mini (\texttt{gpt-5-mini-}\allowbreak\texttt{2025-08-07}), Gemini 2.5 Pro (\texttt{gemini-2.5-pro}), Claude Sonnet 4.5 (\texttt{claude-sonnet-4-5-}\allowbreak\texttt{20250929}), or the complete consensus flow. Direct inference from the same three closed models provides the reference rows in Table~\ref{tab:main-results}. MedGemma-1.5-4B-IT~\citep{sellergren2026medgemma15technicalreport} and HuatuoGPT-o1-7B~\citep{chen2025huatuogpto1} provide external medical references. \textsc{GRPO + DPO} is our matched baseline: it uses the same data, three-state rewards, accepted exact-tie pairs, seed, and training budget as \conrub{}, adding an unweighted DPO term with \(\beta=2\) to vanilla \grpo{}~\citep{rafailov2023dpo,lanchantin2025bridgingofflineonlinereinforcement}.

\paragraph{Data and training.}
Criterion-source variants use four frozen 5,166-prompt manifests with identical prompt order. They use binary scoring, the same seed, and the same three-epoch schedule, so the comparison changes the criterion source while holding the prompt set, reward form, and training budget fixed. The prompts comprise 70\% RaR-Medicine~\citep{gunjal2026rubricsasrewards}, 20\% RubricHub Medical, 5\% RubricHub Chat, and 5\% RubricHub Instruction-Following~\citep{li2026rubrichub}; all 5,166 are unique after normalization. We use the last checkpoint saved during the third epoch for all benchmark evaluations. Each rollout contains 32 prompts and eight responses per prompt (global batch size 256). We use Adam~\citep{kingma2015adam} with learning rate \(3\times10^{-6}\), maximum response length 8,192, and rollout temperature 1.0. The final \conrub{} training run uses one node with eight NVIDIA H200 GPUs.

\paragraph{Evaluation.}
Medical benchmarks are HealthBench-Hard~\citep{arora2025healthbenchevaluatinglargelanguage}, MedXpertQA-Text~\citep{zuo2025medxpertqa}, DiagnosisArena-MCQ~\citep{zhu2026diagnosisarena}, MedMCQA~\citep{pal2022medmcqa}, PubMedQA~\citep{jin2019pubmedqa}, and an aggregate over MMLU medical subjects~\citep{hendrycks2021mmlu}; WritingBench~\citep{wu2025writingbench}, GPQA-Diamond~\citep{rein2024gpqa}, and IFEval prompt-level loose accuracy~\citep{zhou2023instructionfollowingevaluationlargelanguage} assess generalization and capability retention. HealthBench-Hard uses the official GPT-4.1 judge, \texttt{gpt-4.1-2025-04-14}~\citep{openai2025gpt41}. All policy comparisons use one training seed. For \conrub{}, HealthBench-Hard and GPQA-Diamond are averaged over three and five evaluation seeds, respectively; all remaining scores use one fixed evaluation pass. For controlled comparisons, all trained variants are evaluated on the same benchmark examples with identical decoding settings, output limits, and invalid-output handling.

\paragraph{Expert evaluation of criterion quality.}
We evaluate criterion quality on 17 complete matched cases retained from an initial random sample of 24 RaR-Medicine questions. For each question, the final panel from the complete construction pipeline is paired with a single-generator panel containing the same number of criteria and the same mix of criteria covered and not covered by the reference answer. The panels are anonymized, and two medical-domain experts independently evaluate all 136 criteria in each condition for medical validity, atomicity, clinical relevance, and non-redundancy. The supplementary material gives the full protocol.

\subsection{Main Results}

\paragraph{Overall performance.}
Across the open-weight models in Table~\ref{tab:main-results}, \conrub{} ranks first on six of nine benchmarks: HealthBench-Hard, MedXpertQA-Text, DiagnosisArena-MCQ, WritingBench, GPQA-Diamond, and MMLU-Medical. It also obtains the highest medical and generalization macro-averages, 51.76 and 74.27. Relative to Qwen3-4B-Instruct-2507, it improves HealthBench-Hard by 32.68 points and GPQA-Diamond by 17.41 points, and it outperforms the matched \textsc{GRPO + DPO} baseline on all six medical benchmarks. Supplementary Section ``Additional Qwen3-32B Results'' reports a separate Qwen3-32B before-and-after comparison.

\paragraph{HealthBench-Hard comparison.}
Figure~\ref{fig:healthbench-hard-literature} shows a HealthBench-Hard mean of 38.98 and an observed range of 37.94--40.02. The entire range exceeds the 37.30 point estimate of the strongest listed InfiMed-ORBIT setting~\citep{lyu2026clinalign,wang2026infimedorbit}.

\paragraph{Criterion quality.}
Both experts rate 99.3\% of the criteria from the complete construction pipeline as clinically relevant, compared with 86.8\% and 86.0\% for the matched single-generator panels. The corresponding gains are 12.5 and 13.2 percentage points, with 95\% CIs of \([6.6,19.1]\) and \([6.6,20.6]\). Medical validity is near ceiling in both conditions, while atomicity and non-redundancy show no clear difference.

\subsection{Ablation Studies}

Table~\ref{tab:ablations} reports the component-level comparisons. Among the binary-scoring variants, Consensus Rubrics obtain the highest medical average, 49.26 versus 47.51--48.93 for the single-generator conditions. The largest gains are on HealthBench-Hard and GPQA-Diamond, where Consensus exceeds the strongest single-generator result by 6.69 and 8.69 points, respectively; it also leads on WritingBench. The improvement is not uniform, however: its IFEval-Loose score is lower than those of the single-generator variants, leaving a generalization average of 64.69.

Keeping Consensus Rubrics fixed, Three-State scoring raises the medical average by 2.25 points, from 49.26 to 51.51. The gains include 3.69 points on HealthBench-Hard, 3.30 on DiagnosisArena, and 6.80 on PubMedQA, while the three generalization scores change little or decline. Pairwise Advantages change the medical average by only 0.25 points, from 51.51 to 51.76, but raise the generalization average from 62.96 to 74.27, including gains of 3.53 points on GPQA-Diamond and 30.32 on IFEval-Loose. In this comparison, Three-State scoring mainly improves medical benchmarks, whereas Pairwise Advantages raise the generalization average with little change in the medical average.

\subsection{Analysis}

\paragraph{Reward and response length.}
The binary-scoring runs reach higher training rewards than the Three-State runs, 0.780--0.802 versus 0.751--0.753 in Figure~\ref{fig:training-reward-trajectories}(a--b), yet under the matched Consensus setting they obtain a lower medical average, 49.26 versus 51.51 in Table~\ref{tab:ablations}. \conrub{} also produces the shortest responses among the compared variants. This pattern is consistent with reward hacking: with binary scoring, a longer answer can mention content associated with more criteria and accumulate more positive credits, even though some of the added content may be unsupported or hallucinated. By assigning negative credit to incorrect content, Three-State scoring reduces the benefit of indiscriminately expanding the answer.

\paragraph{Reward saturation.}
In the \conrub{} run, the exact-tie rate rises from 2.80\% in epoch 1 to 7.73\% in epoch 3, while reward 1.0 accounts for 64.6\% and 85.2\% of those ties, respectively (Figure~\ref{fig:training-reward-trajectories}c). As training progresses, scalar reward therefore distinguishes fewer responses, leaving more groups with zero vanilla \grpo{} advantages.

\paragraph{Pairwise comparisons in tied groups.}
Of the 847 exact-tie groups, the bidirectional judge accepts 120 preferences and provides nonzero sequence advantages to 90 groups (10.63\%). Pairwise Advantages leave the tail reward nearly unchanged, 0.753 to 0.751, but raise the generalization average from 62.96 to 74.27 while changing the medical average only slightly, from 51.51 to 51.76. Longer and shorter responses each win 60 accepted comparisons, indicating that the pairwise judge does not systematically favor longer or shorter responses. Supplementary Section ``Case Study'' shows a calculation error identified consistently in both candidate orders despite identical criterion profiles and scalar rewards.

\section{Limitations}

The three generators may share the same error, and unanimous support can exclude useful criteria identified by only one or two models. Our source-blind expert study covers 17 matched RaR-Medicine questions; broader evaluation across medical topics and criterion sources would provide a stronger test of the construction pipeline. Rubric construction also relies on several large hosted models, and future work could examine whether smaller open models provide comparable criteria at lower cost.

Pairwise Advantages are restricted to exact final-reward ties. Extending them to near ties would require a calibrated margin that preserves meaningful scalar differences. Their directions also come from a Qwen3-32B judge; agreement across judges and with clinician preferences remains to be studied. Given the compute cost of eight-response on-policy rollouts with an online 32B judge, we limit controlled policy comparisons to one training seed. Repeated runs across seeds and model families remain future work.

\section*{Ethical Statement}

This study uses model-generated criteria and model judges to train medical QA policies. Errors in either can become training signals and reinforce inaccurate or unsafe responses. The medical-domain expert study assesses criterion quality on a limited sample, while the policy evaluation uses research benchmarks; neither establishes safety for diagnosis, treatment, or patient-facing use. Clinical use would require independent validation in the intended setting.

\section{Conclusion}

We introduced \conrub{}, a rubric-guided reinforcement learning approach that connects consensus criterion construction and Three-State scoring with Pairwise Sequence Advantages for exact reward ties. In a source-blind study, two medical experts rated criteria from the consensus construction pipeline as more clinically relevant than matched single-generator criteria, and \conrub{} ranked first on six of nine benchmarks among the evaluated open-weight models. Matched ablations clarify how the feedback path contributes: Three-State scoring mainly improves medical performance, while Pairwise Advantages produce larger generalization gains by recovering an optimization signal in tied groups. Rubric quality is only one part of effective supervision. Criterion judgments must also remain informative through scoring and actionable during policy optimization.

\FloatBarrier
\bibliography{conrubmed2027}

\clearpage
\setcounter{topnumber}{4}
\setcounter{totalnumber}{6}
\setcounter{dbltopnumber}{4}
\renewcommand{\topfraction}{0.95}
\renewcommand{\dbltopfraction}{0.95}
\renewcommand{\textfraction}{0.05}
\renewcommand{\floatpagefraction}{0.80}
\renewcommand{\dblfloatpagefraction}{0.80}
\section*{Supplementary Material}
\appendix

\section{Dataset Construction and Reproducibility}
\label{app:data-repro}

\subsection{\conrub{} Training Dataset and Prompt Provenance}

The run artifacts record model identifiers, retry policies, deterministic
filters, review decisions, criterion-level support, source distribution, and
validation results.

The final \conrub{} training dataset contains 5,166 prompts sampled from
RaR-Medicine and the Medical, Chat, and Instruction-Following subsets of
RubricHub at a fixed 70/20/5/5 ratio~\citep{gunjal2026rubricsasrewards,li2026rubrichub}.
For every prompt, three models independently propose 20 to 30 atomic criteria
from the question and available reference context. The proposals are reviewed,
semantically clustered, and retained under unanimous cross-model support; where
reference-coverage tags are available, agreement on those tags is also required.
After criterion-level review, exact deduplication, and capping, we add two fixed
global control criteria to each prompt. The resulting 59,378 instances
comprise 49,046 consensus-constructed content criteria and 10,332 global
controls.

\subsection{Criterion-Source Ablation Setup}

The criterion-source comparison uses four 5,166-prompt manifests with the same
prompt order. The matched criterion panels come from GPT-5 Mini~\citep{openai2025gpt5mini},
Gemini 2.5 Pro~\citep{comanici2025gemini25pushingfrontier}, Claude Sonnet
4.5~\citep{anthropic2025claudesonnet45}, or the reviewed consensus construction.
All four arms use binary satisfaction and the same three-epoch schedule, so
they differ only in criterion source.

\subsection{Final Dataset Statistics and Labels}
\label{app:training-manifest}

The criterion-scoring and exact-tie comparisons use the full dataset described
above. Table~\ref{tab:data-stats} reports its source allocation and criterion
counts. Each prompt contains 6 to 12 criteria (mean 11.49; median and 90th
percentile 12), with at most ten content criteria and two global controls. The
70/20/5/5 split refers to prompt provenance; all prompt-specific panels use the
consensus construction.

\paragraph{Source processing and label semantics.}
Source-level preprocessing normalizes questions, removes duplicates, and
selects prompts under the fixed quotas; all 5,166 retained prompts are unique
after normalization. The source allocation records where the prompts
originate, while the same three-generator review-and-consensus process produces
criteria for all four source strata. All prompt-specific criteria use
\texttt{binary\_specification=true}. Where construction reference coverage is
defined, \texttt{reference\_judgment=yes/no} records whether the reference
covers the criterion; it is not a reward label. Criteria without a defined
construction reference and global controls use a null value.

\begin{table}[t]
\centering
\small
\setlength{\tabcolsep}{4pt}
\begin{tabular}{lr}
\toprule
Statistic & Value \\
\midrule
RaR-Medicine candidate records & 6,019 \\
Final prompts, unique after normalization & 5,166 \\
RaR-Medicine prompts & 3,616 \\
RubricHub Medical prompts & 1,033 \\
RubricHub Chat prompts & 258 \\
RubricHub Instruction Following & 259 \\
Consensus-constructed content criteria & 49,046 \\
Global control criteria & 10,332 \\
Total criterion instances & 59,378 \\
Mean criteria per prompt & 11.49 \\
Median / 90th percentile / maximum & 12 / 12 / 12 \\
Three-generator support & 100\% \\
\bottomrule
\end{tabular}
\caption{Composition of the final \conrub{} training set. The prompts follow
the 70/20/5/5 source ratio. All content criteria use the consensus construction,
and each prompt has two global controls.}
\label{tab:data-stats}
\end{table}

\paragraph{Benchmark overlap.}
We first checked for normalized exact matches and then retrieved candidate
near-duplicates using lexical and semantic similarity. A total of 12 potential
duplicate items were identified (Table~\ref{tab:benchmark-overlap}).

\begin{table}[t]
\centering
\small
\setlength{\tabcolsep}{3pt}
\begin{tabular}{lrr}
\toprule
Benchmark & Items & Rate \\
\midrule
HealthBench-Hard & 0 & 0.00\% \\
MedXpertQA-Text & 8 & 0.33\% \\
DiagnosisArena-MCQ & 0 & 0.00\% \\
MedMCQA & 4 & 0.10\% \\
PubMedQA & 0 & 0.00\% \\
MMLU-Medical & 0 & 0.00\% \\
WritingBench & 0 & 0.00\% \\
GPQA-Diamond & 0 & 0.00\% \\
IFEval-Loose & 0 & 0.00\% \\
\midrule
Total & 12 & 0.10\% \\
\bottomrule
\end{tabular}
\caption{Potential duplicate items identified after exact-match checking and
candidate-level screening. Rates are computed over unique prompts in each
benchmark.}
\label{tab:benchmark-overlap}
\end{table}

\paragraph{Dataset construction and reward inputs.}
A standalone script builds the dataset, and an independent validator checks the
result. The stored artifacts include the final JSONL, prompt-source lists,
generator and review outputs, consensus-cluster metadata, validation results,
source quotas, and filtering counts. The reward function uses only criterion text:
criterion weights and deterministic-rule routing are inactive, and the length
penalty uses the global rollout limit instead of
\texttt{target\_response\_tokens}. These settings are recorded in the run
manifest.

\subsection{Training Configuration}
\label{app:training-config}

Table~\ref{tab:training-config} gives the configuration of the final Qwen3-4B
run. The Three-State and Pairwise-Advantage ablations use the same settings;
only the feedback component under study changes.

\FloatBarrier
\begin{table}[h]
\centering
\small
\setlength{\tabcolsep}{4pt}
\renewcommand{\arraystretch}{1.03}
\begin{tabular}{@{}P{0.105\textwidth}P{0.33\textwidth}@{}}
\toprule
Setting & Value \\
\midrule
Model & Qwen3-4B-Instruct-2507 \\
\addlinespace[2pt]
Training & Prompts: 5,166 \\
& Epochs: 3 \\
\addlinespace[2pt]
Rollout & Prompts per batch: 32 \\
& Responses per prompt: 8 \\
& Global batch size: 256 \\
& Maximum response length: 8,192 tokens \\
\addlinespace[2pt]
Sampling & Temperature: 1.0 \\
& Top-\(p\): 1.0 \\
& Top-\(k\): \(-1\) \\
& Thinking: disabled \\
\addlinespace[2pt]
Optimizer & Adam \\
& Learning rate: \(3\times10^{-6}\) \\
& Weight decay: 0.1 \\
& Precision: BF16 \\
\addlinespace[2pt]
Judge & Model/service: Qwen3-32B via SGLang \\
& Temperature: 0 \\
& Concurrency: 16 \\
& Retries: up to 8 \\
& Timeout: 120\,s \\
\addlinespace[2pt]
Seed & 1234 \\
\addlinespace[2pt]
Hardware & GPUs: 8 NVIDIA H200 GPUs (143{,}771 MiB each) \\
& CPU: Intel Xeon Platinum 8558 (128 cores available) \\
& System memory: 1.5 TB \\
\addlinespace[2pt]
System & Container OS: Ubuntu 24.04.2 (overlay filesystem) \\
& Linux kernel: 5.10.134 \\
& NVIDIA driver: 550.127.08 \\
& CUDA: 12.9 \\
\addlinespace[2pt]
Software & Python: 3.12.3 \\
& PyTorch: \texttt{2.9.1+cu129} (cuDNN 9.16.0) \\
& Transformers: 4.57.1; Ray: 2.54.0 \\
& SGLang: 0.5.9; FlashAttention: 2.7.4.post1 \\
\addlinespace[2pt]
Compute & Single-node, colocated policy/rollout execution \\
& Policy/rollout: GPUs 0 to 3 (TP 2; 2 GPUs per engine) \\
& Judge: GPUs 4 to 7 (TP 4; port 30012) \\
\bottomrule
\end{tabular}
\caption{Training configuration for the final Qwen3-4B run.}
\label{tab:training-config}
\end{table}

\newpage
\subsection{Models and Evaluation}
\label{app:models-evaluation}

Table~\ref{tab:models-evaluation} distinguishes direct-inference references
from the controlled Qwen3-4B policies and summarizes the reporting protocol
for all nine benchmarks. The external medical references are
MedGemma-1.5-4B-IT and HuatuoGPT-o1-7B~\citep{sellergren2026medgemma15technicalreport,chen2025huatuogpto1};
the closed-source references are the same three models used during criterion
construction~\citep{openai2025gpt5mini,comanici2025gemini25pushingfrontier,anthropic2025claudesonnet45}.

\FloatBarrier
\begin{table*}[t]
\centering
\small
\setlength{\tabcolsep}{5pt}
\renewcommand{\arraystretch}{1.05}
\begin{tabular}{@{}P{0.20\textwidth}P{0.30\textwidth}P{0.44\textwidth}@{}}
\toprule
\multicolumn{3}{@{}l}{\textbf{Models and baselines}} \\
\addlinespace[2pt]
Entry & Role & Identifier or configuration \\
\midrule
GPT-5 Mini & Criterion generator and closed-source reference & \texttt{gpt-5-mini-2025-08-07}; direct API inference \\
Gemini 2.5 Pro & Criterion generator and closed-source reference & \texttt{gemini-2.5-pro}; direct API inference \\
Claude Sonnet 4.5 & Criterion generator and closed-source reference & \texttt{claude-sonnet-4-5-20250929}; direct API inference \\
MedGemma-1.5-4B-IT & External open-weight medical reference & MedGemma-1.5-4B-IT; direct inference \\
HuatuoGPT-o1-7B & External open-weight medical reference & HuatuoGPT-o1-7B; direct inference \\
Qwen3-4B-Instruct-2507 & Controlled initialization & \texttt{Qwen/Qwen3-4B-Instruct-2507}; direct inference before training \\
\textsc{GRPO + DPO} & Trained comparison & Qwen3-4B initialization; Three-State reward with an exact-tie DPO term (\(\beta=2\)) \\
\conrub{} & Full method & Qwen3-4B initialization; Three-State reward and Pairwise Advantages \\
\bottomrule
\end{tabular}

\vspace{7pt}

\begin{tabular}{@{}P{0.22\textwidth}P{0.47\textwidth}P{0.25\textwidth}@{}}
\toprule
\multicolumn{3}{@{}l}{\textbf{Benchmarks}} \\
\addlinespace[2pt]
Benchmark & Reported metric and evaluator & Final \conrub{} reporting \\
\midrule
HealthBench-Hard & Official rubric-judge score using \texttt{gpt-4.1-2025-04-14} & Mean of three evaluation passes \\
MedXpertQA-Text & Official text-task evaluator score & One fixed pass \\
DiagnosisArena-MCQ & Official multiple-choice judge score & One fixed pass \\
MedMCQA & Answer-choice accuracy & One fixed pass \\
PubMedQA & Answer-choice accuracy & One fixed pass \\
MMLU-Medical & Aggregate accuracy over the medical subjects & One fixed pass \\
WritingBench & Official benchmark judge score & One fixed pass \\
GPQA-Diamond & Pass@1 accuracy & Mean of five evaluation passes \\
IFEval-Loose & Prompt-level loose instruction-following accuracy & One fixed pass \\
\bottomrule
\end{tabular}
\caption{Models, baselines, and evaluation protocol used in the main comparison.}
\label{tab:models-evaluation}
\end{table*}
\FloatBarrier

The medical and generalization averages are unweighted means across the six and
three displayed benchmarks, respectively. All policy comparisons use one
training seed. Within each benchmark, all compared models use the same decoding
settings, output limits, examples, and invalid-output handling.

\section{Expert Evaluation of Criterion Quality}
\label{app:criterion-audit}

\paragraph{Design.}
Before annotation, we randomly selected a fixed set of 24 RaR-Medicine
questions, with eight criteria per condition. We assigned one generator to
each question, then paired its panel with a consensus panel of the same size
and the same numbers of positive and negative reference-coverage judgments.
Six sampled questions were absent from the final training set. One more could
not meet the fixed reference-coverage composition. The complete-case sample
therefore contains 17 matched pairs and 136 criteria per condition. No
exclusion depends on an expert label. The single-generator panels come from
Claude, Gemini, and GPT for 4, 6, and 7 questions, respectively.

\paragraph{Blinding and annotation.}
The annotation package shows only the medical question and an anonymous
criterion panel. It omits the condition, generator identity, reference answer,
reference-coverage labels, cross-model support, and source provenance. Two
medical experts independently assign four binary labels to every criterion:
(1) \emph{medical validity}, whether rewarding the criterion would be medically
correct, safe, and non-misleading for the question; (2) \emph{atomicity},
whether it expresses one independently judgeable requirement; (3)
\emph{clinical relevance}, whether satisfying it would materially improve the
answer, including clinically useful optional detail; and (4)
\emph{non-redundancy}, whether it adds a distinct scoring target relative to
the other criteria in the panel. The annotators do not see the condition key
or each other's labels before completing the audit. The archived files
identify them as Expert A and Expert B.

\paragraph{Analysis.}
We leave disagreements unadjudicated and report pass rates separately for each
annotator. For each metric, the effect is the within-question difference
between final consensus and matched single-generator conditions. We compute
percentile 95\% confidence intervals using 10,000 bootstrap replicates at the
question level, stratified by the assigned single generator. All estimates use
the same complete-case sample. A criterion passes ``all four'' only if it
passes all four quality labels. Table~\ref{tab:criterion-agreement} reports raw
agreement and Cohen's \(\kappa\) over all 272 criteria.

\begin{table*}[t]
\centering
\scriptsize
\setlength{\tabcolsep}{3.2pt}
\begin{tabular}{lrrr@{\hspace{6pt}}rrr}
\toprule
& \multicolumn{3}{c}{Expert A} & \multicolumn{3}{c}{Expert B} \\
\cmidrule(lr){2-4}\cmidrule(lr){5-7}
Criterion property
& Single & Consensus & $\Delta$ [95\% CI]
& Single & Consensus & $\Delta$ [95\% CI] \\
\midrule
Medical validity
& 98.5 & 100.0 & $+1.5$ [0.0, 3.7]
& 98.5 & 100.0 & $+1.5$ [0.0, 3.7] \\
Atomicity
& 88.2 & 83.1 & $-5.1$ [$-13.2$, 2.2]
& 89.0 & 86.0 & $-2.9$ [$-10.3$, 3.7] \\
Clinical relevance
& 86.8 & 99.3 & $+12.5$ [6.6, 19.1]
& 86.0 & 99.3 & $+13.2$ [6.6, 20.6] \\
Non-redundancy
& 97.8 & 95.6 & $-2.2$ [$-5.1$, 0.7]
& 93.4 & 90.4 & $-2.9$ [$-8.1$, 2.9] \\
All four properties
& 76.5 & 79.4 & $+2.9$ [$-7.4$, 12.5]
& 72.8 & 78.7 & $+5.9$ [$-4.4$, 15.4] \\
\bottomrule
\end{tabular}
\caption{Expert evaluation of criterion quality. Pass rates (\%) are based on 136 criteria per condition, with panels matched by question, size, and reference coverage. \(\Delta\) is consensus minus single generator in percentage points; brackets show 95\% question-cluster bootstrap confidence intervals. ``All four'' requires all listed properties.}
\label{tab:expert-criterion-evaluation}
\end{table*}

\begin{table}[t]
\centering
\small
\setlength{\tabcolsep}{4pt}
\begin{tabular}{lrr}
\toprule
Dimension & Raw agreement & Cohen's \(\kappa\) \\
\midrule
Medical validity & 0.993 & 0.496 \\
Atomicity & 0.901 & 0.573 \\
Clinical relevance & 0.930 & 0.475 \\
Non-redundancy & 0.930 & 0.357 \\
All four properties & 0.809 & 0.463 \\
\bottomrule
\end{tabular}
\caption{Inter-annotator agreement on the 272 criteria in the matched expert evaluation.}
\label{tab:criterion-agreement}
\end{table}

\paragraph{Interpretation.}
For clinical relevance, both experts rate the final consensus panels higher
than the matched single-generator panels, and both 95\% confidence intervals
exclude zero. The intervals for the other dimensions include zero.

\section{Additional Qwen3-32B Results}
\label{app:qwen32b-results}

We train Qwen3-32B on the same 5,166 prompts using the same feedback
formulation~\citep{yang2025qwen3technicalreport}. Table~\ref{tab:qwen32b-results}
compares the model before and after three epochs. Benchmark
definitions follow their original releases~\citep{arora2025healthbenchevaluatinglargelanguage,zuo2025medxpertqa,zhu2026diagnosisarena,pal2022medmcqa,jin2019pubmedqa,hendrycks2021mmlu,wu2025writingbench,rein2024gpqa,zhou2023instructionfollowingevaluationlargelanguage}.
This is a single before-and-after Qwen3-32B experiment; the controlled
ablations in the main paper use Qwen3-4B.

\begin{table}[h]
\centering
\small
\setlength{\tabcolsep}{5pt}
\begin{tabular}{@{}lrr@{}}
\toprule
Benchmark & Base & After 3 epochs \\
\midrule
HealthBench-Hard & 9.46 & 33.17 \\
MedXpertQA-Text & 22.41 & 17.55 \\
DiagnosisArena-MCQ & 48.60 & 47.20 \\
MedMCQA & 69.04 & 68.59 \\
PubMedQA & 71.60 & 74.40 \\
MMLU-Medical & 85.22 & 86.08 \\
Medical average & 51.06 & 54.50 \\
\midrule
WritingBench & 74.49 & 79.32 \\
GPQA-Diamond & 56.26 & 61.11 \\
IFEval-Loose & 87.25 & 87.99 \\
Generalization average & 72.67 & 76.14 \\
\bottomrule
\end{tabular}
\caption{Single-run Qwen3-32B results before and after three epochs of \conrub{} training on 5,166 prompts. Averages are unweighted across six medical and three generalization benchmarks.}
\label{tab:qwen32b-results}
\end{table}

Six of nine benchmark scores increase after three epochs, while three decline. The
medical macro-average changes from 51.06 to 54.50, and the generalization
macro-average changes from 72.67 to 76.14. The medical results are mixed:
HealthBench-Hard increases from 9.46 to 33.17, while MedXpertQA-Text,
DiagnosisArena-MCQ, and MedMCQA decline.

\section{Reward and Optimization Details}
\label{app:reward-details}

Three-State scoring assigns \(+1\), \(0\), and \(-1\) to \correct{},
\missing{}, and \wrong{}, respectively, and averages these credits across the
criteria for a response. Binary scoring uses \(1\) for a satisfied criterion
and \(0\) otherwise.

For response length \(L\), rollout limit \(M\), buffer \(B\), and weight \(w_\ell\), both pointwise reward variants use DAPO-style soft overlength shaping~\citep{yu2025dapo}:
\[
p_{\mathrm{len}}(L)=
\begin{cases}
0, & L\le M-B,\\
-w_\ell\dfrac{L-(M-B)}{B}, & M-B<L\le M,\\
-w_\ell, & L>M.
\end{cases}
\]
We set \(M=8{,}192\), \(B=2{,}048\), and \(w_\ell=1\); the penalty begins after 6,144 tokens and reaches \(-1\) at the rollout limit.

Let \(\rho_{i,t}(\theta)\) be the token-level importance ratio between the current and old policies. The optimizer-facing token advantage defined in the main paper enters a response-level, asymmetrically clipped objective~\citep{shao2024deepseekmathpushinglimitsmathematical,yu2025dapo}:
\begin{equation}
\begin{aligned}
\mathcal L_{\mathrm{GRPO}}(\theta)
&= -\frac{1}{k}\sum_{i=1}^{k}\frac{1}{T_i}
\sum_{t=1}^{T_i}
\min\Bigl(
\rho_{i,t}(\theta)A_{i,t}^{\mathrm{train}},\\[-2pt]
&\quad\operatorname{clip}\!\left(
\rho_{i,t}(\theta),1-\epsilon_{\mathrm{low}},1+\epsilon_{\mathrm{high}}
\right)A_{i,t}^{\mathrm{train}}
\Bigr).
\end{aligned}
\end{equation}
We set \(\epsilon_{\mathrm{low}}=0.20\) and
\(\epsilon_{\mathrm{high}}=0.28\), clip the gradient norm at 1.0, and use zero
KL and entropy coefficients. Here
\(A_{i,t}^{\mathrm{train}}=\widetilde A_i\): it equals
\(A_i^{\mathrm{GRPO}}\) for non-tied groups and \(d_i\) for eligible exact
final-reward ties.

\paragraph{\textsc{GRPO + DPO} comparison.}
The \textsc{GRPO + DPO} run uses the same Qwen3-4B initialization, 5,166
training prompts, Three-State reward, exact-tie eligibility rule,
bidirectional judging procedure, and seed as \conrub{}. It is an independent
on-policy run, so accepted pairs are generated from its own rollouts. We retain
native \grpo{} advantages and add an unweighted DPO term for each accepted pair.

Let \(s_\theta(y)\) be the mean response-token log probability under the
current policy, and let \(s_{\mathrm{old}}(y)\) be the corresponding score
under the rollout policy. Define
\(\Delta_\theta=s_\theta(y^+)-s_\theta(y^-)\) and
\(\Delta_{\mathrm{old}}=s_{\mathrm{old}}(y^+)-s_{\mathrm{old}}(y^-)\).
For the accepted-pair set \(\mathcal P\), the added loss is
\begin{equation}
\mathcal L_{\mathrm{DPO}}(\theta)
=-\frac{1}{B_{\mathrm{global}}}
\sum_{(y^+,y^-)\in\mathcal P}
\log\sigma\!\left(\beta[
\Delta_\theta-\Delta_{\mathrm{old}}]\right).
\end{equation}
We use \(\beta=2\) and optimize
\(\mathcal L_{\mathrm{GRPO}}+\mathcal L_{\mathrm{DPO}}\). The outer
normalizer is the global batch size, not the number of valid pairs. A step with
no accepted pair has zero DPO contribution.

\section{Training Records and Derived Metrics}
\label{app:trace-fields}

The traces link sample, group, and pair records. Sample records contain run,
prompt, and response identifiers, optional raw text, token counts,
truncation and removal status, the ordered criteria, criterion-level verdicts
and credits, reward components, and judge metadata. Group records add final
rewards, equality and variance fields, sequence advantages, and the source of
each advantage. Pair records retain both candidate orders, parsed and mapped
decisions, chosen and rejected identities, response lengths, pair weight, and
objective type. Training metadata stores the coefficient and advantage source
applied to each sample.
Native \grpo{} advantages can be reconstructed from the recorded rewards and
normalization settings.

We compute each derived metric only when its required fields are present. We
omit groups with a removed, failed, or nonfinite candidate from pairwise
comparison and exclude those candidates from valid-only clinical metrics.
Cross-run comparisons match candidates by prompt, rollout batch, and candidate
position. We report the exact-tie group rate,
order-consistent-pair rate, nonzero pairwise-advantage coverage, the share of
accepted pairs won by the longer response, and a check that non-tied groups
retain their reconstructed \grpo{} advantages.

\paragraph{Exact reward ties.}
Each exact-tie group contains eight responses with identical final scalar rewards under the logger's equality test. The complete run has 144 exact ties in epoch 1, 305 in epoch 2, and 398 in epoch 3, out of 5,152 groups per epoch. For the reward histogram, values are rounded to one decimal. Of the 847 ties, 784 (92.56\%) fall in bins from 0.8 to 1.0, including 682 (80.52\%) in the 1.0 bin.

Pairwise comparisons assign nonzero sequence advantages to 26, 27, and 37 tied groups in the three epochs. Across the run, 90 groups and 240 responses receive such advantages in 77 updates. Under vanilla \grpo{}, every response in these exact-tie groups would have zero group-relative advantage. The bidirectional order check accepts 120 of 3,388 candidate pairs (3.54\%). For the remaining pairs, the judge returns \emph{same} 2,923 times, changes its decision after the response order is swapped 314 times, and produces 31 invalid outputs.

\begin{figure*}[t]
\centering
\includegraphics[width=\textwidth]{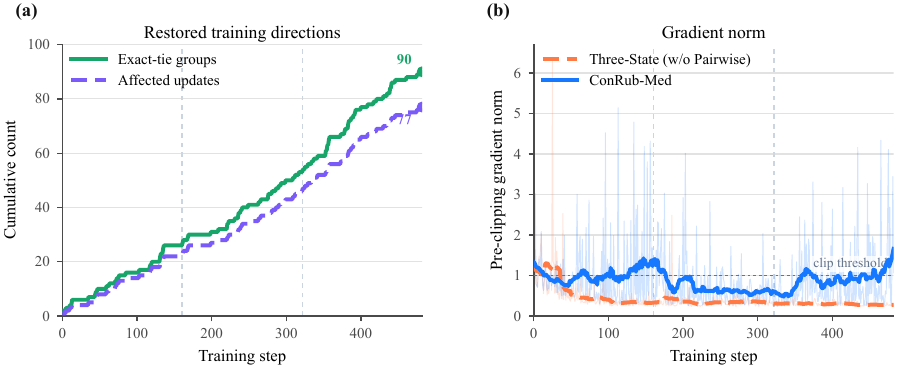}
\caption{Optimization signals during matched \conrub{} and Three-State training. (a) Cumulative numbers of exact-tie groups assigned a nonzero direction and updates containing at least one such group. The 90 groups are distributed across 77 of 483 updates. (b) Gradient norm before clipping; the horizontal dotted line marks the clipping threshold at 1.0. Thin curves show individual updates, thick curves show 21-update rolling means, and vertical dotted lines mark epoch boundaries.}
\label{fig:optimizer-signal-dynamics}
\end{figure*}

\paragraph{Optimization signal recovery.}
Pairwise Advantages restore group-relative training directions throughout optimization. Across three epochs, 90 of 847 exact-tie groups (10.63\%) receive nonzero sequence advantages. These groups make up 0.58\% of all 15,456 rollout groups, but they are distributed across 77 of 483 updates (15.94\%), as shown in Figure~\ref{fig:optimizer-signal-dynamics}(a). In every affected update, at least one exact-tie group that would contribute no group-relative policy-gradient signal under vanilla \grpo{} instead supplies a nonzero training direction.

Across steps 0 to 338, the GPT-5 Mini, Gemini 2.5 Pro, Claude Sonnet 4.5, and Consensus binary runs contain 729, 635, 1,001, and 518 exact-tie groups out of 10,848 groups per run. Their rounded reward 1.0 bins contain 634, 532, 868, and 405 groups, respectively.

\begin{figure*}[t]
\centering
\includegraphics[width=\textwidth]{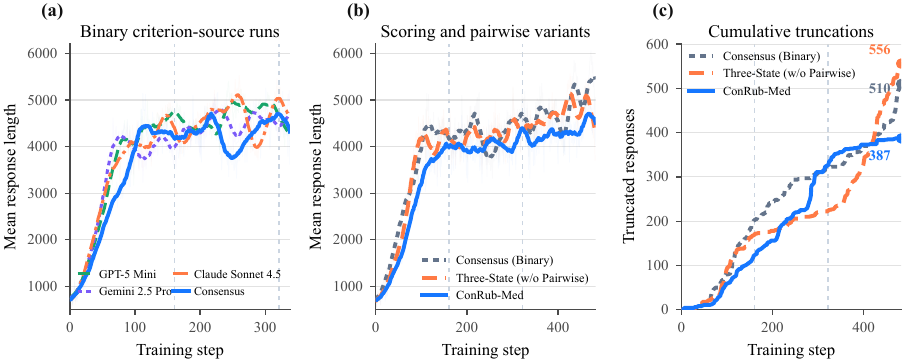}
\caption{Response length and truncation during training. (a) Mean response length for four binary scoring runs, each using criteria from a different source, over roughly 340 updates. (b) Mean response length for Consensus with binary scoring, Three-State without Pairwise Advantages, and \conrub{}. (c) Cumulative number of responses truncated at the rollout limit for the same three runs. Thin curves show individual updates, and thick curves show rolling means with a window of 21 updates. Vertical dotted lines mark epoch boundaries.}
\label{fig:training-efficiency-analysis}
\end{figure*}

Across steps 0 to 338, the mean response lengths in the final 10\% of updates are 4,635 tokens for GPT-5 Mini, 4,573 for Gemini 2.5 Pro, 4,860 for Claude Sonnet 4.5, and 4,551 for Consensus with binary scoring, compared with 4,247 for \conrub{}. Across the complete runs, the corresponding tail means are 5,223 for Consensus with binary scoring, 4,738 for Three-State without Pairwise Advantages, and 4,481 for \conrub{}. The three runs produce 510, 556, and 387 truncated responses. \conrub{} has the shortest tail responses and the fewest truncations. The pairwise judge selects the longer response in 60 of 120 accepted comparisons.

\begin{figure*}[t]
\centering
\includegraphics[width=\textwidth]{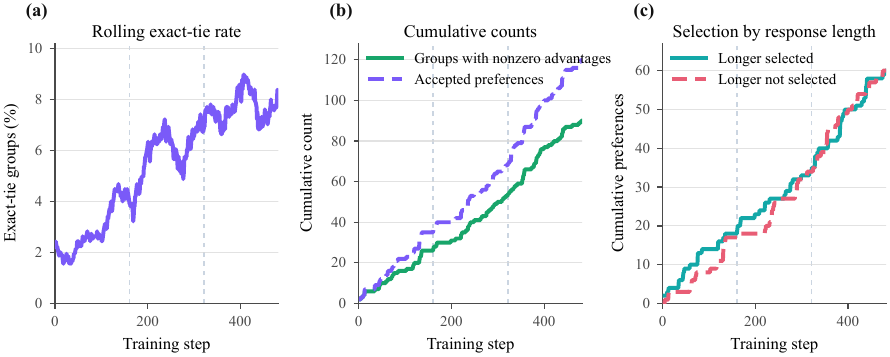}
\caption{Pairwise comparisons for exact reward ties over 483 updates. (a) Exact-tie rate averaged over 32 updates. (b) Cumulative numbers of tied groups assigned nonzero sequence advantages and accepted preferences. (c) Accepted preferences separated by whether the longer response is selected. Vertical dotted lines mark epoch boundaries.}
\label{fig:pairwise-signal-trajectories}
\end{figure*}

\paragraph{Update-level optimization.}
Figure~\ref{fig:optimizer-signal-dynamics}(a) shows that Pairwise Advantages restore training directions throughout all three epochs rather than in a single short interval. The gradient norm before clipping is higher for \conrub{} in 432 of 483 updates; its median and mean are 0.571 and 0.881, compared with 0.313 and 0.409 for Three-State.

\begin{figure*}[t]
\centering
\includegraphics[width=\textwidth]{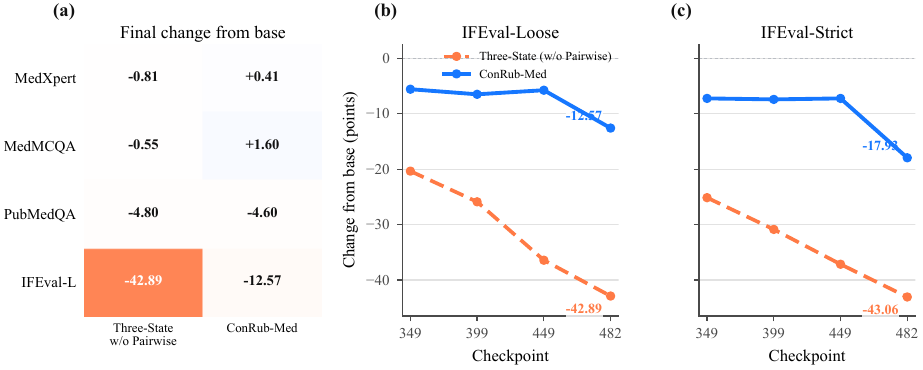}
\caption{Capability retention during medical specialization. (a) Final change from the shared base on all four main-table benchmarks for which Three-State finishes below its starting score. (b,c) Change from the same base at four saved checkpoints under IFEval loose and strict scoring. Positive values indicate improvement over the base and negative values indicate decline.}
\label{fig:capability-retention}
\end{figure*}

\paragraph{Capability retention across benchmarks.}
At the final checkpoint, Three-State finishes below the shared base on four of the nine main-table benchmarks: MedXpertQA-Text, MedMCQA, PubMedQA, and IFEval-Loose, with changes of \(-0.81\), \(-0.55\), \(-4.80\), and \(-42.89\) points. Figure~\ref{fig:capability-retention}(a) reports this complete set. With Pairwise Advantages, the corresponding changes are \(+0.41\), \(+1.60\), \(-4.60\), and \(-12.57\) points. Pairwise Advantages recover the small MedXpertQA-Text and MedMCQA declines, leave the final PubMedQA score nearly unchanged, and reduce the IFEval-Loose decline by 30.32 points.

\paragraph{Instruction-following retention.}
Relative to Three-State, the IFEval gains at steps 349, 399, 449, and 482 are 14.79, 19.41, 30.69, and 30.32 points under loose scoring, and 17.92, 23.47, 29.94, and 25.13 points under strict scoring. At the final checkpoint, Pairwise Advantages reduce the observed decline from the shared base by 70.7\% under loose scoring and 58.4\% under strict scoring. IFEval-Loose accounts for 30.32 of the 33.95 summed benchmark-point increase across the three general benchmarks (89.3\%), while the medical macro-average changes by only 0.25 points. The agreement of loose and strict scores across all four checkpoints shows a sustained instruction-following retention effect rather than an isolated endpoint difference.

\begin{figure*}[t]
\centering
\includegraphics[width=\textwidth]{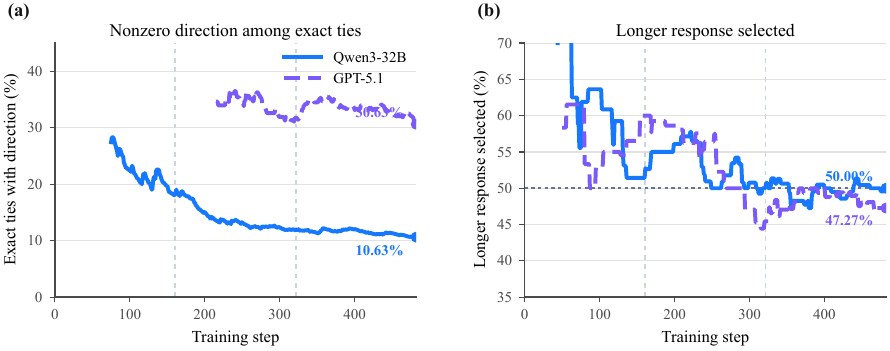}
\caption{Sensitivity to the pairwise judge under the same 4B policy setup. (a) Share of exact ties assigned a nonzero training direction, shown after each run encounters 50 exact ties. (b) Cumulative fraction of accepted comparisons that select the longer response, shown after ten accepted comparisons. The pairwise judges are Qwen3-32B and GPT-5.1.}
\label{fig:pairwise-judge-sensitivity}
\end{figure*}

\paragraph{Comparison of pairwise judges.}
Under the same 4B policy setup, GPT-5.1~\citep{openai2025gpt51} assigns a
nonzero training direction to 30.63\% of exact ties, compared with 10.63\% for
Qwen3-32B. GPT-5.1 selects the longer response in 47.27\% of accepted
comparisons; Qwen3-32B does so in 50.00\%.

\section{Judge Prompts and Output Formats}

Section~\ref{app:prompt-templates} reproduces the construction and judge
templates together with their output interfaces. The Chinese sample-review
prompt used in the run is shown there in English translation.
The run artifacts record the parser rules, retry policy, model versions, and
decoding settings.

\subsection{Three-State Criterion Judge}

The Three-State judge receives the prompt, one candidate response, and the
ordered prompt-specific criteria. It labels a criterion \correct{} when the
response addresses it with factually correct content, \missing{} when the
response says nothing useful about it, and \wrong{} only when the response
contains a factual error, fabrication, contradiction, reversal, or
substitution concerning that point. It returns one JSON label per criterion.

\subsection{Bidirectional Pairwise Judge}

We invoke the pairwise judge only for complete eight-response groups with
finite, exactly equal rewards and no removed or failed response. A shuffle
seeded by the run seed and group identity forms four disjoint pairs. For each
pair, the judge receives the question, prompt-specific criteria, and candidates
A and B without seeing their scalar rewards or the reference answer. The judge
bases its comparison primarily on the criteria. Clear medical errors, unsafe
claims, or contradictions outside the panel may also distinguish the
responses. It returns A, B, or \emph{same}, then repeats the comparison with the
candidates swapped. We accept a preference only when both mapped outputs select
the same response; otherwise both pair coefficients are zero.

\begin{algorithm*}[t]
\caption{Pairwise sequence advantages for exact final-reward ties}
\label{alg:pairwise}
\begin{algorithmic}[1]
\REQUIRE Question \(q\), response group \(Y\), criteria \(C(q)\)
\STATE Judge every \((y_i,c_j)\) and compute final rewards \(R_i\)
\STATE Compute group-relative advantages \(A_i^{\mathrm{GRPO}}\)
\IF{the group is incomplete, contains a failed response, has a nonfinite reward, or is not exactly tied}
  \RETURN \(R_i\) and \(A_i^{\mathrm{GRPO}}\)
\ENDIF
\STATE Shuffle the eight indices and form four disjoint pairs
\FOR{each pair \((y_a,y_b)\in\mathcal P_g\)}
  \STATE Compare \((A=y_a,B=y_b)\) and \((A=y_b,B=y_a)\)
  \IF{both mapped outputs select the same response}
    \STATE Assign winner \(+1\) and loser \(-1\)
  \ELSE
    \STATE Assign both responses \(0\)
  \ENDIF
\ENDFOR
\RETURN \(R_i\) and sequence advantages \(\widetilde{A}_i\)
\end{algorithmic}
\end{algorithm*}

\clearpage
\onecolumn
\section{Case Study}
\label{app:exact-tie-case}

\begin{tcolorbox}[
  enhanced,
  colback=white,
  colframe=SuppBlack,
  boxrule=0.8pt,
  arc=1.8mm,
  left=3mm,
  right=3mm,
  top=2.5mm,
  bottom=2.5mm,
  title={\strut When equal scalar scores hide a unit-conversion error},
  colbacktitle=SuppBlack,
  coltitle=white,
  fonttitle=\bfseries\large
]
\small
\begin{tcolorbox}[
  colback=SuppLightGray,
  colframe=SuppLightGray,
  boxrule=0pt,
  arc=1mm,
  left=2mm,
  right=2mm,
  top=1.3mm,
  bottom=1.3mm
]
\begin{tabular}{@{}P{0.22\textwidth}P{0.24\textwidth}P{0.22\textwidth}P{0.24\textwidth}@{}}
\textbf{Source} & Final \conrub{} training trace
& \textbf{Final reward} & \(1.0\) for both responses \\
\textbf{Criterion profile} & \(7/7\) \correct{} for both
& \textbf{Pairwise decision} & Response 1 in both orders \\
\end{tabular}
\end{tcolorbox}

Both responses receive the same scalar reward and criterion profile, even though
one contains a unit-conversion error. The excerpts retain only the calculation
relevant to the pairwise decision; \([\ldots]\) marks omitted context.

\begin{tcolorbox}[
  colback=SuppCodeGray,
  colframe=SuppDarkGray,
  boxrule=0.45pt,
  arc=1mm,
  title={Question and target criterion},
  colbacktitle=SuppDarkGray,
  coltitle=white,
  fonttitle=\bfseries
]
The total neutron dose rate at 1 ft is
\[
\dot{H}_n=\left(\frac{S_n}{4\pi r^2}\right)k,
\qquad
S_n=2.0\times10^5\ \mathrm{n/sec},
\qquad
k=\frac{2.5\ \mathrm{mrem/hr}}
        {20\ \mathrm{n\,sec^{-1}\,cm^{-2}}}.
\]
One criterion explicitly requires the final value \(2.14\) mrem/hr.
\end{tcolorbox}

\noindent
\begin{minipage}[t]{0.485\textwidth}
\begin{tcolorbox}[
  enhanced,
  colback=SuppPaperGray,
  colframe=SuppDarkGray,
  boxrule=0.55pt,
  arc=1.2mm,
  title={Response 1: candidate 3},
  colbacktitle=SuppDarkGray,
  coltitle=white,
  fonttitle=\bfseries,
  equal height group=CaseResponses
]
\[
[\ldots]
\]
\[
\phi\approx17.09\ \mathrm{n\,cm^{-2}\,s^{-1}}.
\]
\[
\dot{H}_n=\phi\times k
=17.09\times0.125
=2.136\ \mathrm{mrem/hr}.
\]
\[
\boxed{\dot{H}_n\approx2.14\ \mathrm{mrem/hr}}
\]
\[
[\ldots]
\]
\end{tcolorbox}
\end{minipage}
\hfill
\begin{minipage}[t]{0.485\textwidth}
\begin{tcolorbox}[
  enhanced,
  colback=SuppPanelGray,
  colframe=SuppDarkGray,
  boxrule=0.55pt,
  arc=1.2mm,
  title={Response 2: candidate 7},
  colbacktitle=SuppDarkGray,
  coltitle=white,
  fonttitle=\bfseries,
  equal height group=CaseResponses
]
\[
1\ \mathrm{n\,cm^{-2}\,s^{-1}}
=\frac{2.5}{20}
=0.125\ \mathrm{mrem/hr}.
\]
\[
[\ldots]
\]
\[
\dot{H}_n=\Phi\times k
=17.09\times0.125
=2.136\ \mathrm{mrem/s}.
\]
The response then converts this value to mrem/hr:
\[
2.136\ \mathrm{mrem/s}\times3600
=7{,}690\ \mathrm{mrem/hr}.
\]
\[
[\ldots]
\]
\end{tcolorbox}
\end{minipage}

\begin{tcolorbox}[
  colback=SuppCodeGray,
  colframe=SuppDarkGray,
  boxrule=0.45pt,
  arc=1mm,
  title={Bidirectional order check},
  colbacktitle=SuppDarkGray,
  coltitle=white,
  fonttitle=\bfseries
]
\centering
\setlength{\tabcolsep}{10pt}
\begin{tabular}{@{}lllll@{}}
\toprule
Call & Candidate A & Candidate B & Judge output & Mapped winner \\
\midrule
Forward & Response 1 & Response 2 & A & Response 1 \\
Swapped & Response 2 & Response 1 & B & Response 1 \\
\bottomrule
\end{tabular}
\end{tcolorbox}

\begin{tcolorbox}[
  colback=SuppLightGray,
  colframe=SuppBlack,
  boxrule=0.6pt,
  arc=1.2mm,
  title={Why the pairwise decision differs},
  colbacktitle=SuppBlack,
  coltitle=white,
  fonttitle=\bfseries
]
The conversion factor \(k\) already maps flux in
\(\mathrm{n\,cm^{-2}\,s^{-1}}\) to dose rate in mrem/hr; Response 2 introduces
an extra factor of 3,600 by treating the product as mrem/s. The pairwise judge
therefore selects Response 1 in both orders, although the scalar criterion
scores are identical. The accepted preference assigns \(+1\) to Response 1 and
\(-1\) to Response 2.
\end{tcolorbox}
\end{tcolorbox}

\clearpage

\section{Prompt Templates}
\label{app:prompt-templates}

This section reproduces the instructions used for data construction and reward
judging. Braced fields denote runtime substitutions, and numbered criteria are
expanded to the length of the current panel. Line wrapping and typographic
dashes are normalized for the paper, but the wording and output constraints are
unchanged.

\subsection{Data Construction}

The four templates below define the LLM-assisted consensus construction used
for the prompt-specific panels across all source strata. Generator inputs use
the question and available reference context. After semantic grouping, a
deterministic exporter retains only clusters supported by all three generators
and, where reference-coverage tags are available, requires those tags to agree.
The clustering prompt's minimum group size of two is therefore a proposal rule,
not the final retention rule.

\begin{PromptFrame}{Rubric generation: user prompt}
\begin{PromptText}
You are an expert rubric writer for medical questions. Given a question and its reference answer, generate a set of binary (yes/no) evaluation criteria.

Requirements:
1. Each criterion is answerable strictly as YES or NO.
2. Be SPECIFIC and CONCRETE - bad: "The response is comprehensive"; good: "The response explains that tryptophan is converted to serotonin and then melatonin."
3. Cover the question from MULTIPLE ANGLES. Think broadly about what a thorough, helpful answer would include:
   - Core facts and reasoning (diagnosis, mechanism, pathophysiology)
   - Supporting details (lab values, imaging, clinical signs, epidemiology)
   - Differentials or alternative explanations when relevant
   - Management: treatment options, contraindications, dosage/duration
   - Risk factors, complications, prognosis
   - Practical guidance: when to seek care, next steps, follow-up
   - Safety: red flags, warnings, special populations
   - Clarity: logical structure, appropriate level of detail
4. Don't repeat the same point in different words. Each criterion should check a DISTINCT piece of information.
5. Include some criteria the reference answer does NOT satisfy (judgment=no) - these represent stretch goals or extra depth.

Generate 20-30 criteria. For each, judge whether the reference answer satisfies it.

Output a JSON array. Each element has exactly 2 keys:
- "criterion": one sentence starting with "The response..."
- "reference_judgment": "yes" or "no"

<question>
{question}
</question>

<reference_answer>
{reference_answer}
</reference_answer>

Output ONLY the JSON array, no other text.
\end{PromptText}
\end{PromptFrame}

\begin{PromptFrame}{Medical sample review prompt used in the run (English translation)}
\begin{PromptText}
You are a rigorous medical data-quality reviewer and scoring expert. Complete the following two tasks.

TASK 1: CONTRADICTION DETECTION

1. Read the Question and Reference Answer carefully.
2. Read every Rubric criterion, which was independently generated by multiple AI models.
3. Determine whether the Reference Answer naturally satisfies most positive Rubrics (criteria with reference_judgment=yes).
4. Determine whether the Reference Answer violates any criterion that should be avoided.
5. MOST IMPORTANT: Use your medical knowledge to determine whether the logic or conclusions encouraged by the Rubrics contain serious medical factual errors.

Contradiction rule:
- Serious medical factual error, logical contradiction, or severe conflict between the Reference Answer and Rubrics => REJECT
- Minor wording differences or omission of optional content do not constitute a conflict => PASS

TASK 2: ALIGNMENT SCORE (only if Task 1 is PASS)

Check whether the Reference Answer satisfies each Rubric and assign a score from 0 to 10:
- Satisfies all core criteria (correct diagnosis and key facts) = start from 8
- Adds depth and completeness = 9-10
- Misses a core criterion = reduce directly below 7
- Contains a factual error = 5 or below

DATA

<question>
{question}
</question>

<reference_answer>
{reference_answer}
</reference_answer>

<rubrics>
{rubrics_text}
</rubrics>

OUTPUT FORMAT

Output only one JSON object and no other text.

If a serious contradiction exists:
{"verdict": "REJECT", "reason": "brief reason for the conflict"}

If no contradiction exists:
{"verdict": "PASS", "score": N, "satisfied_count": X, "total_count": Y, "issues": "brief description of unsatisfied core criteria; empty string if all are satisfied"}
\end{PromptText}
\end{PromptFrame}

Samples with \texttt{verdict=REJECT} or a score below 8 are removed before
semantic grouping.

\begin{PromptFrame}{Semantic grouping: user prompt}
\begin{PromptText}
You are given rubric criteria generated by multiple AI models for evaluating the same medical response. Your task is to group criteria that evaluate the SAME thing (same semantic meaning, even if worded differently).

Each criterion has an ID like "A1", "B3" etc. where the letter is the model and the number is the criterion index within that model.

Output a JSON array of groups. Each group is an array of IDs that share the same semantic meaning.
- Only group criteria that truly evaluate the same fact/aspect (not just vaguely related)
- A criterion can only appear in one group
- Criteria that are unique to one model (no match) should NOT appear in any group
- Minimum group size: 2

Example output:
[["A1","B3","C2"], ["A5","C7"], ["B1","C1"]]

Here are the criteria:

{criteria_text}

Output ONLY the JSON array, no other text.
\end{PromptText}
\end{PromptFrame}

\begin{PromptFrame}{Criterion audit: system prompt}
\begin{PromptText}
You are a meticulous auditor of medical QA rubrics.

Task:
Classify each rubric criterion. You are auditing the rubric text itself, not
grading a model answer.

Important distinction:
- A valid medical safety criterion can contain words like "avoid",
  "contraindicated", "not indicated", or "not appropriate". This is NOT a bad
  negative rubric if the criterion asks the response to state a substantive
  medical safety fact.
  Example: "The response states that bupropion is contraindicated in patients
  with eating disorders" => safety_constraint_medical, keep.
- A problematic negative rubric rewards absence, silence, or non-mention, or
  is double-negation ambiguous.
  Example: "The response does not mention antibiotics" or "The response should
  not include X" => problematic_negative_absence, rewrite/drop/manual_review.
- A criterion with reference_judgment="no" is not automatically negative. It
  often means the reference answer did not cover a valid supplementary point.

Rubric types:
1. direct_positive_medical
2. safety_constraint_medical
3. exclusion_or_wrong_option_explanation
4. format_or_communication
5. supplementary_but_valid
6. problematic_negative_absence
7. ambiguous_or_non_atomic
8. non_medical_or_low_value

Keep decisions:
- keep: usable as-is.
- rewrite: medically useful but should be canonicalized into a clearer positive
  criterion.
- drop: low-value/problematic enough to remove.
- manual_review: uncertain or needs human medical review.

Return ONLY valid JSON. Do not use markdown fences.
\end{PromptText}
\end{PromptFrame}

\begin{PromptFrame}{Criterion audit: user template}
\begin{PromptText}
Audit the rubrics for one medical QA sample.

Question:
{question}

Reference answer:
{reference_answer}

Rubrics to audit:
{rubric_block}

Return JSON with this exact shape:
{
  "items": [
    {
      "id": "r0",
      "rubric_type": "direct_positive_medical | safety_constraint_medical | exclusion_or_wrong_option_explanation | format_or_communication | supplementary_but_valid | problematic_negative_absence | ambiguous_or_non_atomic | non_medical_or_low_value",
      "is_problematic_negative": false,
      "negative_kind": "none | valid_medical_safety_negation | valid_exclusion_reasoning | absence_reward | double_negation_ambiguous | reference_uncovered_not_negative | other",
      "keep_decision": "keep | rewrite | drop | manual_review",
      "canonical_positive_rewrite": "",
      "confidence": 0.0,
      "reason": "short reason in Chinese"
    }
  ],
  "record_notes": "short Chinese note"
}

Rules:
- Include exactly one item for every input rubric id.
- If a rubric says an answer should mention that a treatment is contraindicated,
  not indicated, inappropriate, or should be avoided for safety reasons, mark it
  as safety_constraint_medical, not problematic_negative_absence.
- Mark is_problematic_negative=true only when the rubric's success condition is
  absence/silence/non-mention, or the wording is so double-negative that a judge
  could reward the wrong behavior.
- If reference_judgment is "no" but the criterion is medically valid, set
  negative_kind="reference_uncovered_not_negative".
\end{PromptText}
\end{PromptFrame}

The audit keeps \texttt{keep} items and replaces \texttt{rewrite} items with
\texttt{canonical\_positive\_rewrite}; \texttt{drop} and
\texttt{manual\_review} items are excluded. If the upstream API rejects the
clinical context, a fallback sends the same rubric-level schema and rules while
omitting the question and reference answer.

\subsection{Criterion Judges}

The scoring calls receive the question, one candidate response, and the ordered
criteria. The output is a prompt-constrained JSON array rather than an
API-enforced JSON schema.

\begin{PromptFrame}{Binary criterion judge: system prompt}
\begin{PromptText}
You are a strict, careful medical-domain evaluator. You judge a candidate response against a list of criteria. For EACH criterion answer whether the response satisfies that criterion.
  - "yes": the response actually states or clearly implies the required content, and that content is medically/factually correct.
  - "no": the response does not address the criterion, is too vague/irrelevant to count, contradicts it, fabricates information, or gives medically/factually wrong information about it.
Be literal: related background is not enough for yes. If the answer is missing, unsupported, ambiguous, or wrong on the criterion, mark no.
\end{PromptText}
\end{PromptFrame}

\begin{PromptFrame}{Binary criterion judge: user template}
\begin{PromptText}
QUESTION:
{question}

CANDIDATE RESPONSE:
{response}

CRITERIA (judge each independently):
1. {criterion_1}
...
N. {criterion_N}

For each criterion in order, assign one of: "yes", "no".
Respond with ONLY a JSON array of {N} strings, in the same order as the criteria. Example: ["yes","no","yes"]. Output nothing else.
\end{PromptText}
\end{PromptFrame}

\begin{PromptFrame}{Three-state criterion judge: system prompt}
\begin{PromptText}
You are a strict, careful medical-domain evaluator. You judge a candidate response against a list of criteria. For EACH criterion you assign one of three labels:
  - "correct": the response actually addresses that criterion AND the content it gives for it is factually right (no fabrication, no contradiction, no medical error).
  - "wrong": the response gives factually WRONG, fabricated, self-contradictory, or medically erroneous information about that point. Judge this by the information the response conveys about that point regardless of whether it "addresses" it head-on: an error reached directly (asserting something false), by reversal (naming the opposite of the right answer), or by substitution (putting a wrong value/item in place of the right one) are all "wrong". The point of the split from "missing": conveying nothing useful about a point is "missing" (neutral); conveying something wrong about it is "wrong" (penalised).
  - "missing": the response does not address that criterion at all (says nothing relevant to it).
Be literal: "correct" requires the response to actually state or clearly imply the required content and to be right about it. Do NOT mark "correct" for content that is merely related. Do NOT mark "wrong" merely because the wording is loose, imprecise, or overly general but not actually incorrect - that is still "missing" or "correct". Reserve "wrong" for genuine factual error/fabrication/contradiction on that point.
\end{PromptText}
\end{PromptFrame}

\begin{PromptFrame}{Three-state criterion judge: user template}
\begin{PromptText}
QUESTION:
{question}

CANDIDATE RESPONSE:
{response}

CRITERIA (judge each independently):
1. {criterion_1}
...
N. {criterion_N}

For each criterion in order, assign one of: "correct", "wrong", "missing".
Respond with ONLY a JSON array of {N} strings, in the same order as the criteria. Example: ["correct","wrong","missing"]. Output nothing else.
\end{PromptText}
\end{PromptFrame}

\subsection{Bidirectional Pairwise Judge}

The same user template is called twice with A and B exchanged. A preference is
accepted only when both mapped winners identify the same underlying response.

\begin{PromptFrame}{Pairwise judge: system prompt}
\begin{PromptText}
You are a strict medical-domain evaluator comparing two candidate responses to the same question under the same evaluation criteria. For each criterion you may judge 'A', 'B', or 'same' (neither clearly better on that criterion). Then give an overall winner which MUST be one of 'A', 'B', 'same'.

Rules:
- Judge each listed criterion by medical correctness, not by length or style. A longer answer is NOT better unless it is more correct.
- 'same' means the two responses are roughly equal in quality on that criterion (including both wrong in the same way).
- 'A'/'B' means that candidate is clearly better on that criterion (more correct, fewer errors).
- The overall winner need not be unanimous across criteria; weigh medical correctness most.
- For the overall winner, use the criteria as the primary anchors, but clear medical errors, unsafe claims, or contradictions outside them may also distinguish the responses. Do not use uncertain outside-criterion claims to break a tie.

Output ONLY a JSON object, nothing else, e.g.:
{"winner":"A","criterion_comparisons":["A","same","B"],"confidence":"high","reason":"short audit text"}
\end{PromptText}
\end{PromptFrame}

\begin{PromptFrame}{Pairwise judge: user template}
\begin{PromptText}
QUESTION:
{question}

CANDIDATE A:
{response_A}

CANDIDATE B:
{response_B}

CRITERIA (compare A vs B on each):
1. {criterion_1}
...
N. {criterion_N}

For each criterion output 'A', 'B', or 'same', then an overall winner in {'A','B','same'}. Respond with ONLY the JSON object.
\end{PromptText}
\end{PromptFrame}

The parser requires \texttt{winner} to be one of \texttt{A}, \texttt{B}, or
\texttt{same}. Criterion-level comparisons, confidence, and the short reason
are retained for trace auditing.

\clearpage
\twocolumn

\end{document}